\documentclass[10pt,journal]{IEEEtranTCOM}

\usepackage[english]{babel}
\usepackage[usenames]{color}
\usepackage[cp1250]{inputenc}
\usepackage{amsfonts}
\usepackage{amsthm}
\usepackage{graphicx}
\usepackage{epsfig}
\usepackage{mathrsfs}
\usepackage{amsmath}
\usepackage{algorithm}
\usepackage{algorithmic}

\theoremstyle{plain}

\begin{document}
\newcommand{\bea}{\begin{eqnarray}}
\newcommand{\eea}{\end{eqnarray}}
\newcommand{\be}{\begin{equation}}
\newcommand{\ee}{\end{equation}}
\newcommand{\beas}{\begin{eqnarray*}}
\newcommand{\eeas}{\end{eqnarray*}}
\newcommand{\bs}{\backslash}
\newcommand{\bc}{\begin{center}}
\newcommand{\ec}{\end{center}}
\def\SC {\mathscr{C}}

\title{SGD momentum optimizer with\\ with step estimation by online parabola model}
\author{\IEEEauthorblockN{Jarek Duda}\\
\IEEEauthorblockA{Jagiellonian University,
Golebia 24, 31-007 Krakow, Poland,
Email: \emph{dudajar@gmail.com}}}
\maketitle

\begin{abstract}
In stochastic gradient descent, especially for neural network training, there are currently dominating first order methods: not modeling local distance to minimum. This information required for optimal step size is provided by second order methods, however, they have many difficulties, starting with full Hessian having square of dimension number of coefficients.

This article proposes a minimal step from successful first order momentum method toward second order: online parabola modelling in just a single direction: normalized $\hat{v}$ from momentum method. It is done by estimating linear trend of gradients $\vec{g}=\nabla F(\vec{\theta})$ in $\hat{v}$ direction: such that $g(\vec{\theta}_\bot+\theta\hat{v})\approx \lambda (\theta -p)$ for $\theta = \vec{\theta}\cdot \hat{v}$, $g= \vec{g}\cdot \hat{v}$, $\vec{\theta}_\bot=\vec{\theta}-\theta\hat{v}$. Using linear regression, $\lambda$, $p$ are MSE estimated by just updating four averages (of $g$, $\theta$, $g\theta$, $\theta^2$) in the considered direction. Exponential moving averages allow here for inexpensive online estimation, weakening contribution of the old gradients. Controlling sign of curvature $\lambda$, we can repel from saddles in contrast to attraction in standard Newton method. In the remaining directions: not considered in second order model, we can simultaneously perform e.g. gradient descent. There is also discussed its learning rate approximation as $\mu=\sigma_\theta / \sigma_g$, allowing e.g. for adaptive SGD - with learning rate separately optimized (2nd order) for each parameter.
\end{abstract}

\section{Introduction}

In many optimization scenairios like neural network trainig, we search for a local minimum of objective/loss function $F(\,\vec{\theta}\,)$ of parameters $\vec{\theta}\in\mathbb{R}^D$, which number is often in millions. The real function is usually unknown, only modelled based on a size $n$ dataset: $F(\,\vec{\theta}\,)=\frac{1}{n}\sum_i F_i(\,\vec{\theta}\,)$. Due to its large size, there is often used SGD (stochastic gradient descent)~\cite{sgd} philosophy: dataset is split into minibatches used to calculate succeeding stochastic gradients $\vec{g}^{\,t}=\textrm{grad}(\,\vec{\theta}^{\,t})$, which can be imagined as noisy (approximate) gradients:
$$\nabla_\theta F(\,\vec{\theta}^{\,t}\,)\approx
\vec{g}^{\,t}=\frac{1}{|\textrm{minibatch}|}\sum_{i\in\textrm{minibatch}} \nabla_\theta F_i(\,\vec{\theta}^{\,t}\,) $$

Efficient training especially of deep neural network requires extraction and exploitation of statistical trends from such noisy gradients: calculated on subsets of samples. Their e.g. exponential moving averaging in momentum method~\cite{momentum} tries to estimate the real gradient of minimized objective function and use it for gradient descent. However, there remains a difficult problem of choosing the step size for such descent.

\begin{figure}[b!]
    \centering
        \includegraphics[width=8.5cm]{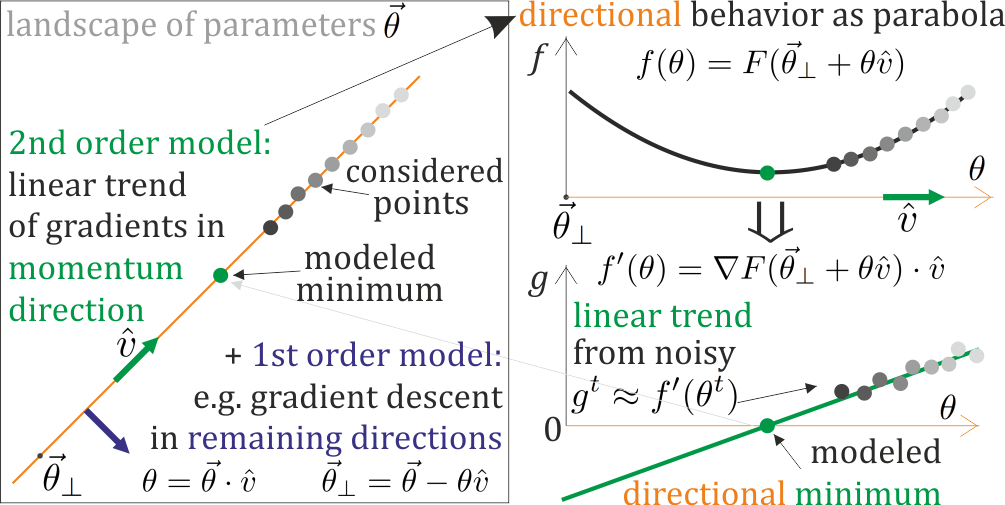}
        \caption{General diagram of the considered approach. Left: we perform momentum method to online choose direction $\hat{v}$ for 2nd order model, in the remaining $\theta_\bot$ directions we can e.g. simultaneously perform gradient descent. Right: in $\hat{v}$ direction we search for linear trend of gradients using linear regression, choose step size accordingly to this model - e.g. proportionally to distance to minimum of modelled parabola. Grayness of considered points represents their fading weights in used exponential moving averages.}
        \label{intr}
\end{figure}

\begin{algorithm}[b!]
\normalsize{
\caption{\textbf{OGR1d}()\quad\{basic $d=1$ online grad. regr.\}}
\begin{algorithmic}
\STATE initialize() \qquad \COMMENT{and choose hyperparameters}
\STATE warmup()   \qquad \COMMENT{find initial direction $\hat{v}$ and averages avg}
\REPEAT
\STATE $\vec{g}\leftarrow $grad($\vec{\theta}$) \qquad\quad  $\vec{v}\leftarrow \gamma \vec{v} + \vec{g}$\qquad\quad \COMMENT{momentum}
\STATE upd\_avg(avg, $\hat{v})$  \qquad\COMMENT {update $g$, $\theta$, $g\theta$, $\theta^2$ averages}
\STATE step(avg, $\hat{v}$)\quad\ \COMMENT{get parabola from avg, make $\vec{\theta}$ step}
\STATE $ \hat{v}\leftarrow\vec{v}/\|\vec{v}\|_2$    \qquad\qquad \COMMENT{update modeled direction}
\UNTIL step limit or convergence condition
\end{algorithmic}
}
\label{alg1}
\end{algorithm}

For example in a plateau we should greatly increase the step, simultaneously being careful not to jump over a valley. We will use linear trend of gradients to estimate position of bottom of such valley as parabola: where the linear trend of derivatives in the considered direction intersects zero, as visualized in Fig. \ref{intr} with basic pseudocode as Algorithm 1.

Linear trend can be estimated in MSE (mean squared error) optimal way with linear regression, which requires four averages: here of $g$, $\theta$, $g\theta$, $\theta^2$ for the linear relation between position $\theta$ and the first derivative $g$. There will be used exponential moving averages for inexpensive online update and to reduce reliance on the old gradients - they have exponentially weakening weights.

There will be also discussed $\textrm{corr}(g,\theta)=1$ approximation which should resolve necessity of clipping, additionally simplifying the method to $\sigma_\theta / \sigma_g$ learning rate, e.g. for adaptive SGD: inexpensive online optimization of learning rate separately for each parameter.\\

Linear trend of gradients is a second order model. Generally, higher than first order methods are often imagined as impractically costly, for example full Hessian would need $O(D^2)$ coefficients. We focus here on the opposite end of cost spectrum - only model parabola in a single $(d=1)$ direction (parameterized by just 2 additional coefficients), for example in direction $\hat{v}$ found by the momentum method: suggesting increased local activity, hence deserving a higher order model. Calculated gradient, beside updating momentum and parabola model, can be also simultaneously used for e.g. gradient descent in the remaining directions.

For low cost it would be preferred to estimate second order behavior from the stochastic gradients only. It is done for example in L-BFGS~\cite{L-BFGS}. However, it estimates inverted Hessian from just a few recent noisy gradients: leading to stability issues and having relatively large cost of processing all these large gradients in each step. In  contrast, thanks to working on updated averages, this processing cost becomes practically negligible in the proposed online gradient linear regression approach. We should also get a better estimation as instead of just a few recent noisy gradients, here we are using all of them with exponentially weakening weights in updated averages.

Another addressed here problem of many second order methods is attraction to saddles e.g. by standard Newton method, which handling can lead to large improvements as shown e.g. in saddle-free Newton (SFN) method article~\cite{SFN}. This repairment requires to control the signs of curvatures, what is relatively difficult and costly. In the presented approach it becomes  simple as we need to control it in only a single direction.

A natural extension is analogously performing such second order modelling in a few dimensions, which was the original approach~\cite{me}. The purpose of this separate article is focusing on the simplest case for introduction and  better understanding of the basics.
\section{1D case with linear regression of derivatives}
We would like to estimate second order behavior from a sequence of gradients: first order derivatives, which linear behavior corresponds to second order derivative. A basic approach is finite differences~\cite{HFN}, for Hessian $H$: $$H\vec{v}=\lim_{\epsilon\to 0} (\nabla f(\vec{\theta}+\epsilon \vec{v})-\nabla f(\,\vec{\theta}\,))/\epsilon$$ However, we have noisy gradients here, hence we need to use much more than two of them to estimate linear trend from their statistics. A standard tool for extraction of linear trend is least squares linear regression: optimal in MSE way. Additionally, it is very convenient due to working on averages: we can replace it with exponential moving averages for online estimation and to weaken contribution of less certain old gradients. Let us now focus on $d=1$ dimensional case, its general $d$-dimensional expansion is discussed in \cite{me}.

\subsection{1D static case - parabola approximation}
Let us start with 1D case - static parabola model as:
$$f(\theta)=h+\frac{1}{2}\lambda(\theta-p)^2\qquad\qquad f'(\theta)=\lambda (\theta-p) $$
and MSE optimizing its parameters for $(\theta^t,g^t)$ sequence:
$$\arg\min_{\lambda,p}\, \sum_t w^t (g^t-\lambda(\theta^t-p))^2\quad\textrm{for some weights } (w^t)$$
For parabola and $t=1,\ldots,T$ times we can choose uniform weights $w^t=1/T$. Later we will use exponential moving average - reducing weights of old noisy gradients, seeing such parabola as only local approximation. The necessary $\partial_p=\partial_\lambda=0$ condition (neglecting $\lambda=0$ case) becomes:
$$ \sum_t w^t (g^t-\lambda(\theta^t-p)) = 0 =\sum_{t}w^t (\theta^t-p)(g^t -\lambda \theta^t +\lambda p) $$
$$\overline{g}-\lambda\overline{\theta}+s\lambda p = 0 = \overline{\theta g}-\lambda\overline{\theta^2}+2\lambda p\overline{\theta}-p\overline{g}-s\lambda p^2$$
for averages:
$$s=\sum_t w^t\qquad\quad\overline{\theta}=\sum_t w^t \theta^t\qquad \quad \overline{g}=\sum_t w^t g^t$$
\be  \overline{g\theta }=\sum_t w^t g^t\theta^t \qquad\qquad \overline{\theta^2}=\sum_t w^t (\theta^t)^2 \label{fa}\ee
Their solution (least squares linear regression) is:
\be \lambda=\frac{s\,\overline{g\theta}-\overline{g}\,\overline{\theta}}
{s\,\overline{\theta^2}-\overline{\theta}^2 }\qquad \qquad
 p=\frac{\lambda\overline{\theta}-\overline{g}}{s\lambda}=\frac{\overline{\theta^2}\,\overline{g} - \overline{\theta}\,\overline{g \theta }}
{\overline{\theta}\,\overline{g}-s\,\overline{g\theta }}\label{reg}\ee

Observe that $\lambda$ estimator is $(g,\theta)$ covariance divided by variance of $\theta$ (positive if not all values are equal).
\subsection{Online update by exponential moving average}
The optimized function is rather not a parabola, should be only locally approximated this way. To focus on local situation we can reduce weights of the old gradients. It is very convenient to use exponential moving averages $w^t\propto \beta^{-t}$ for some $\beta\in(0,1)$ for this purpose as they can be inexpensively updated to get online estimation of local situation. Starting with all 0 values for $t=0$, for $t\geq 1$ we get update rules:

$$\overline{\theta}^t=\beta\, \overline{\theta}^{t-1}+(1-\beta)\,\theta^t=(1-\beta)\sum_{t'=1}^t \beta^{t-t'}\, \theta^{t'}$$
$$\overline{g}^t=\beta\, \overline{g}^{t-1}+(1-\beta)\,g^t\qquad\qquad \overline{g\theta }^t=\beta\, \overline{g\theta }^{t-1}+(1-\beta)\,g^t\theta^t $$
$$\overline{\theta^2}^t=\beta\, \overline{\theta^2}^{t-1}+(1-\beta)\,(\theta^t)^2$$
\be s^t=(1-\beta)\sum_{t'=1}^t \beta^{t-t'}=1-\beta^t=\beta s^{t-1}+(1-\beta) \label{avg} \ee
The $s^t$ is analogous e.g. to bias in ADAM method~\cite{adam}, in later training it can be assumed as just $s=1$.

\subsection{1D linear regression based optimizer}
Linear regression requires values in at least two points, hence there is needed at least one step (better a few) warmup - evolving using e.g. gradient descent, simultaneously updating averages (\ref{avg}), starting from initial $\overline{\theta}^0=\overline{g}^0=\overline{g\theta}^0=\overline{\theta^2}^0=s^0=0$. Then we can start using linear model for derivative: $f'(\theta)\approx \lambda (\theta-p)$, using updated parameters from (\ref{reg}) regression formula.

Getting $\lambda>0$ curvature, the parabola has minimum in $p$, the modeled optimal position would be $\theta=p$. However, as we do not have a complete confidence in such model, would like to work in online setting, a safer step is $\theta\leftarrow \theta+\alpha (p-\theta)$ for some $\alpha\in(0,1]$ parameter describing trust in the model, which generally can vary e.g. depending on estimated uncertainty of parameters. Natural gradient method corresponds to $\alpha=1$ complete trust. Lower $\alpha>0$ would still give exponential decrease of distance from a fixed minimum.

Getting $\lambda<0$, parabola has maximum instead - second order method does not longer suggest a position of minimum. Such directions are relatively rare in neural network training~\cite{hess}, especially focusing on the steepest descent direction here. In many second order methods curvature signs are ignored - attracting to saddles e.g. in standard Newton method. Controlling sign of $\lambda$ here, we can handle these cases - there are two basic approaches~\cite{HFN}: switch to gradient descent there, or reverse sign of step from second order method.

There are also $\lambda \approx 0$ cases, which are problematic as corresponding to very far predicted extremum $p$ in (\ref{reg}) - require some clipping of step size. Such situation can correspond to plateau, or to inflection point: switching between convex and concave behavior. For plateaus we need to use large steps.

While it leaves opportunities for improvements, for simplicity we can for example use SFN-like step: just reversing sign for $\lambda<0$ directions. Applied clipping prevents $\lambda\approx 0$ cases, alternatively we could e.g. replace sign with $\tanh$:
\be \theta^{t+1}=\theta^t+\alpha\,\textrm{sign}(\lambda^t)\, C(p^t-\theta^t) \label{step}\ee
with example of clipping: $C(x)=\textrm{sign}(x)\,\min(|x|,\epsilon)$.
\begin{algorithm}[b!]
\normalsize{
\caption{\textbf{upd\_avg}(avg,\ $\hat{v}$)\quad  \{of avg $=(\overline{\theta},\overline{g},\overline{g\theta},\overline{\theta^2},s)\}$}
\label{dec1}
\begin{algorithmic}
\STATE $\theta \leftarrow \vec{\theta}\cdot \hat{v}$\qquad\qquad \COMMENT{$\vec{\theta},\vec{g}$ are global variables here}
\STATE $g\leftarrow \vec{g}\cdot \hat{v}$

\COMMENT {Formula (\ref{avg}): update 4 averages and bias $s$}
$$\textrm{avg} \leftarrow \beta\,\textrm{avg}+(1-\beta)\,(\theta, g, g\theta,\theta^2,1)$$
\end{algorithmic}
}
\end{algorithm}
\begin{algorithm}[b!]
\normalsize{
\caption{\textbf{step}(avg,\ $\hat{v}$)\qquad \{proper parameter update\}}
\label{dec1}
\begin{algorithmic}
\STATE $(\overline{\theta},\overline{g},\overline{g\theta},\overline{\theta^2},s)=$ avg \qquad \qquad $g\leftarrow \vec{g}\cdot \hat{v}$

\COMMENT {Calculate trend (\ref{reg}) of $g(\vec{\theta}_\bot+\theta\hat{v})\approx \lambda (\theta -p)$:}
$$\lambda\leftarrow\frac{s\,\overline{g\theta}-\overline{g}\,\overline{\theta}}
{s\,\overline{\theta^2}-\overline{\theta}^2 }\qquad \qquad
 p\leftarrow\frac{\lambda\overline{\theta}-\overline{g}}{s\lambda}=\frac{\overline{\theta^2}\,\overline{g} - \overline{\theta}\,\overline{g \theta }}
{\overline{\theta}\,\overline{g}-s\,\overline{g\theta }}$$

\COMMENT{Step along $\vec{v}$, clipping: $C(x)=\textrm{sign}(x)\,\min(|x|,\epsilon)$}
 $$\vec{\theta}\leftarrow\vec{\theta}+\alpha\, \textrm{sign}(\lambda)\, C(p-\theta)\,\hat{v}\qquad\textrm{\{or grad. desc. for $\lambda<0$\}}$$

\COMMENT{Optional gradient descent in remaining directions:}
$$\vec{\theta}\leftarrow \vec{\theta} -\eta (\vec{g}-g\hat{v}) $$
\end{algorithmic}
}
\end{algorithm}

\begin{algorithm}[b!]
\normalsize{
\caption{\textbf{initialize}()\quad \{and choose hyperparameters\}}
\label{dec1}
\begin{algorithmic}
\STATE $\vec{\theta}\in \mathbb{R}^D$ \qquad \COMMENT{initial parameters}
\STATE $\alpha\in (0,1]$ \quad \COMMENT{step size: confidence in parabola model}
\STATE $\beta\in (0,1)$ \qquad \COMMENT{forgetting rate for linear regression}
\STATE $\gamma\in (0,1)$ \qquad \COMMENT{rate for momentum choosing direction}
\STATE $\epsilon>0$ \qquad \COMMENT{for clipping - handling $\lambda\approx 0$ situations}
\STATE $\eta>0$ \qquad \COMMENT{for neglected directions gradient descent}
\STATE $m\in\mathbb{N}$\qquad \COMMENT{number of steps for warmup and stages}
\STATE avg = (0,0,0,0,0)\qquad \COMMENT {$(\overline{g},\overline{\theta},\overline{g\theta},\overline{\theta^2},s)$ averages}
\STATE $\vec{v}=0\in \mathbb{R}^D$
\end{algorithmic}
}
\end{algorithm}

\begin{algorithm}[b!]
\normalsize{
\caption{\textbf{warmup}()\quad\{initial direction and avg\}}
\label{dec1}
\begin{algorithmic}
\STATE
\COMMENT {Initial direction and averages from momentum method:}
\FOR {$i=1$ to $m$}
\STATE $\vec{g}\leftarrow $grad($\vec{\theta}$) \qquad\qquad $\vec{v}\leftarrow \gamma \vec{v} + \vec{g}$ \quad\qquad $\vec{\theta}\leftarrow \vec{\theta}- \eta \vec{v}$
\ENDFOR
\STATE $\hat{v}\leftarrow\vec{v}/\|\vec{v}\|_2$ \qquad \COMMENT{normalize and use to find averages:}
\FOR {$i=1$ to $m$}
\STATE $\vec{g}\leftarrow $grad($\vec{\theta}$) \qquad\qquad $\vec{v}\leftarrow \gamma \vec{v} + \vec{g}$ \quad\qquad $\vec{\theta}\leftarrow \vec{\theta}- \eta \vec{v}$
\STATE upd\_avg(avg, $\hat{v})$\qquad \qquad \COMMENT{find initial averages}
\ENDFOR
\end{algorithmic}
}
\end{algorithm}

\section{Momentum with online parabola methods}
Having above 1D approach we can use it to model behavior of our function as locally parabola $f(\theta)=F(\vec{\theta}_\bot+\theta\hat{v})$ in $d=1$ dimensional affine space $\{\vec{\theta}+\theta \hat{v}:\theta\in\mathbb{R}\}$ of $\mathbb{R}^D$ space of parameters, still performing first order e.g. gradient descent in the remaining directions.

There is a freedom of choosing this emphasized direction $\hat{v}$, but for better use of such additional cost of higher order model we should choose a locally more promising direction - for example pointed by momentum method. Wanting a few $d$-dimensional promising local subspace instead, we could obtain them e.g. from online-PCA~\cite{OPCA} of recent gradients.

\subsection{Common functions and basic OGR1d}
Algorithms 2, 3, 4, 5 contain common functions, used e.g. in basic $d=1$ dimensional OGR (online gradient regression) as Algorithm 1:
\begin{itemize}
  \item upd\_avg(avg,\ $\hat{v}$) updates all averages (packed into avg vector) based on current position $\vec{\theta}$, gradient $\vec{g}$ and considered direction $\hat{v}$,
  \item step(avg,\ $\hat{v}$) finds parameters of linear trend of derivatives in direction $\hat{v}$ and use them to perform step in this direction. It also optionally performs first order e.g. gradient descent in the remaining directions,
  \item initialize() chooses initial $\vec{\theta}$, hyperparameters,  sets averages and momentum to zero,
  \item warmup() uses $m$ steps of momentum method to choose initial direction $\hat{v}$ and averages avg.
\end{itemize}

Then the basic approach is presented as Algorithm 1: just regularly (online) update the $\hat{v}$ direction of second order model accordingly to momentum method. However, such modification of $\hat{v}$ assumes that averages remain the same in the new direction - effectively rotating the second order model. Such rotation might be problematic, should be performed much slower than updating the averages $(\beta<\gamma<1)$.

The next two subsections suggest ways to improve it: safer approach updating averages simultaneously for the old and new direction, and less expensive shifting center of rotation for updates of $\hat{v}$.

\subsection{Safe variant: updating averages for old and new direction}
Algorithm 6 suggests a safe solution for modification of modelled direction $\hat{v}$ by simultaneously updating two sets of averages: for the previous direction (avgw for $\hat{w}$) used for the proper step, and for the new one (avg for $\hat{v}$). After $m$ steps it switches $(\hat{w}\leftarrow \hat{v})$ to the new direction and starts building from zero averages for the next switch.

\begin{algorithm}[b!]
\normalsize{
\caption{\textbf{OGR1ds}()\quad\{safe $d=1$ online grad. regress.\}}
\label{dec1}
\begin{algorithmic}
\STATE initialize() \qquad \COMMENT{and choose hyperparameters}
\STATE warmup()   \qquad \COMMENT{find initial direction $\hat{v}$ and averages avg}
\REPEAT
\STATE $\hat{w}\leftarrow \hat{v}$\qquad\qquad\COMMENT {previous direction}
\STATE $ \hat{v}\leftarrow\vec{v}/\|\vec{v}\|_2$\qquad\COMMENT {new direction}
\STATE avgw $\leftarrow$ avg\qquad\qquad  avg$\leftarrow (0,0,0,0,0)$
\FOR {$i=1$ to $m$}
\STATE $\vec{g}\leftarrow $grad($\vec{\theta}$) \qquad \quad $\vec{v}\leftarrow \gamma \vec{v} + \vec{g}$\qquad \COMMENT{momentum}
\STATE upd\_avg(avgw, $\hat{w})$  \quad\COMMENT {update for previous dir.}
\STATE upd\_avg(avg, $\hat{v})$  \qquad\COMMENT {update for new direction}
\STATE step(avgw, $\hat{w}$)\qquad\COMMENT {step using previous direction}
\ENDFOR
\UNTIL step limit or convergence condition
\end{algorithmic}
}
\end{algorithm}

\subsection{Faster option: shifing rotation center}
Update of $\hat{v}$ in Algorithm 1 effective rotates second order model around $\vec{c}=0\in\mathbb{R}^D$, which is usually far from the current position $\vec{\theta}$, hence such rotation can essentially damage the model. Such rotation is much safer if shifting its center of rotation closer, e.g. to $\vec{c}=\vec{\theta}$. For this purpose, instead of operating on $F(\vec{\theta})$ function, we can work on $F(\vec{\theta}+\vec{c})$, what does not change gradients - only shifts their positions.

We can periodically update this center $\vec{c}$ to current position, shifting representation: replacing projection $\theta$ with $\theta-\Delta \theta$ for $\Delta \theta=\vec{\theta}\cdot \hat{v}$. This shift requires to modify 3 of  averages:
\begin{itemize}
  \item $\overline{\theta^2}$ transforms to $\overline{(\theta-\Delta\theta)^2}=\overline{\theta^2} -2\,\overline{\theta}\,\Delta\theta+(\Delta\theta)^2$
  \item $\overline{g\theta}$ transforms to $\overline{g(\theta-\Delta\theta})=\overline{g\theta}-\overline{g}\Delta\theta$
  \item $\overline{\theta}$ transforms to $\overline{\theta}-\Delta \theta$.
\end{itemize}
Algorithm 7 is example of such modification of Algorithm 1, in practice we can shift the center in every step by using $m=1$.
\begin{algorithm}[b!]
\normalsize{
\caption{\textbf{OGR1dc}()\quad\{centered rotations\}}
\label{dec1}
\begin{algorithmic}
\STATE initialize() \qquad \COMMENT{and choose hyperparameters}
\STATE warmup()   \quad\ \COMMENT{find initial direction $\hat{v}$ and averages avg}
\STATE $\vec{c}\leftarrow 0\in\mathbb{R}^D$ \qquad \COMMENT{center of rotation}
\REPEAT
\STATE
\COMMENT{Update direction $\hat{v}$ and center of rotation $\vec{c}$\,:}
\STATE $\Delta\theta \leftarrow \vec{\theta}\cdot \hat{v} \qquad\qquad\qquad \hat{v}\leftarrow \vec{v}/\|\vec{v}\|_2 $
\STATE $\overline{\theta^2} \leftarrow \overline{\theta^2} -2\,\overline{\theta}\,\Delta\theta+(\Delta\theta)^2$
\STATE $\overline{g\theta}\leftarrow \overline{g\theta}-\overline{g}\,\Delta\theta \qquad\qquad\qquad \overline{\theta}\leftarrow \overline{\theta}-\Delta \theta$
\STATE $\vec{c}\leftarrow \vec{c}+\vec{\theta}\qquad\qquad\qquad \vec{\theta}\leftarrow 0$
\FOR {$i=1$ to $m$}
\STATE $\vec{g}\leftarrow $grad($\vec{\theta}+\vec{c}$) \qquad  $\vec{v}\leftarrow \gamma \vec{v} + \vec{g}$\qquad \COMMENT{momentum}
\STATE upd\_avg(avg, $\hat{v})$  \qquad\COMMENT {update for both directions}
\STATE step(avg, $\hat{v}$)\qquad\qquad\COMMENT{step for $\vec{\theta}$}
\ENDFOR
\UNTIL step limit or convergence condition
\end{algorithmic}
}
\end{algorithm}
\newpage
\section{Learning rate estimation as $\sigma_\theta/\sigma_g$}
\begin{figure}[b!]
    \centering
        \includegraphics[width=8.5cm]{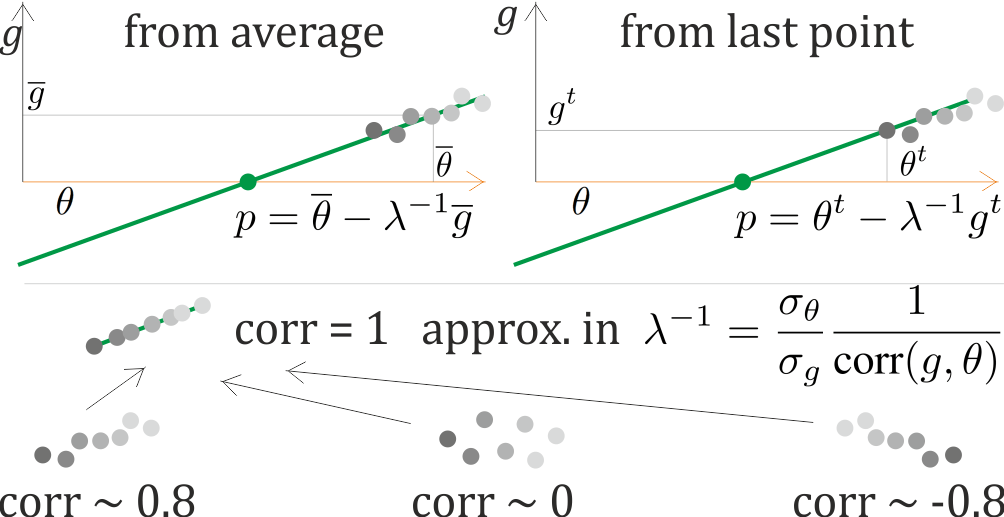}
        \caption{Top: to evaluate position of minimum $p$ in the considered direction, we find linear trend of gradients $g$ in this direction, and ask where this linear trend intersects $g=0$. Left: the derived formula (\ref{reg}) makes step from the averaged position and gradient $(\overline{\theta},\overline{g})$. Right: we can alternatively use the same learning rate as slope $\lambda^{-1} =\frac{\sigma_\theta}{\sigma_g} \frac{1}{\textrm{corr}(g,\theta)}$, but starting from a different point, like the most recent $(\theta^t,g^t)$ instead. Bottom: the $\lambda^{-1}$ formula has intuitive division of dispersion of $\theta$ and of $g$, both are nonnegative, but it also has problematic correlation in denominator - which can go down to 0, and further to negative values for negative curvature directions. Increasing noise of gradients, this correlation would decrease, incorrectly suggesting larger learning rate - to prevent these problems and for simplicity, we can e.g. use $\textrm{corr}(g,\theta)=1$ approximation, using $\eta=\sigma_\theta/\sigma_g$ learning rate, or more generally $\alpha \eta$ for some positive constant $\alpha$.}
        \label{lr}
\end{figure}

Linear regression of gradients has lead to (\ref{reg}) formula for estimated curvature $\lambda$ and position of extremum $p$. Let us rewrite it for $\lambda^{-1}$ and $s=1$ sum of weights in the used averages:
\be \lambda^{-1}=\frac{\overline{\theta^2}-\overline{\theta}^2 }
{\overline{g\theta}-\overline{g}\,\overline{\theta}}\qquad \qquad
 p=\overline{\theta}-\lambda^{-1}\overline{g}\label{reg1}\ee

This $p$ formula is similar to gradient descent using $\lambda^{-1}$ learning rate, but starting from averaged position and using averaged gradient. Wanting to use more standard: last position instead, we can do it maintaining the $\lambda^{-1}$ slope if using the last gradient instead of averaged, as visualized in Fig. \ref{lr}:
\be p=\theta^t - \lambda^{-1}\, g^t \ee

Observe that $\lambda^{-1}$  in (\ref{reg1}) is (weighted) variance of $\theta$, divided by covariance of $g$, $\theta$. Let us define their (weighted) correlation in standard way:
\be\textrm{corr}(g,\theta)=\frac{\overline{g\theta}-\overline{g}\,\overline{\theta}}{\sigma_g\, \sigma_\theta}\in[-1,1] \ee
$$\textrm{for}\quad \sigma_{\theta}=\sqrt{\overline{\theta^2}-\overline{\theta}^2}\qquad \sigma_{g}=\sqrt{\overline{g^2}-\overline{g}^2}$$
where standard $|\textrm{corr}(g,\theta)|\leq 1$ comes from Cauchy-Schwartz inequality $|\langle u,v\rangle|\leq \|u\| \|v\|$, also applying to the discussed weighted case (by using $\langle u,v\rangle=\sum_i w_i u_i v_i$).

It allows to write $\lambda^{-1}$ as:
\be \lambda^{-1} =\frac{\eta}{\textrm{corr}(g,\theta)}\qquad\textrm{for}\qquad  \eta=\frac{\sigma_\theta}{\sigma_g} \ee
Thanks to $|\textrm{corr}(g,\theta)|\leq 1$, we get lower bound for $\lambda^{-1}$ by using practical approximation $\textrm{corr}(g,\theta)=1$ instead as in Fig. \ref{lr}. This way using $\eta=\sigma_\theta/\sigma_g$ learning rate instead, it is always positive: step direction can only point downhill also for negative curvature (repelling saddles), and we have removed looking problematic increase of step size with growing noise. To include such noise dependence, also expressing incomplete trust in parabola model (the real function is not parabola), we could e.g. try to tune additional hyperparmeter $\alpha>0$ and use proper  step:

\be\theta^{t+1}=\theta^t-\alpha\, \eta^t\, g^t\ee

\subsection{Adaptive SGD: coordinate-wise optimized learning rate}

\begin{algorithm}[b!]
\normalsize{
\caption{\textbf{Adaptive SGD}(): \small{coordinate-wise learning rate}}
\label{aSGD}
\begin{algorithmic}
\STATE $\vec{\theta}\in \mathbb{R}^D$ \qquad\quad \COMMENT{initial parameters}
\STATE $\alpha>0$ \qquad\quad\ \ \COMMENT{confidence in parabola model, $\sim 0.5$}
\STATE $\beta\in (0,1)$ \quad\ \ \COMMENT{forgetting rate in averages, $\sim 0.9, 0.99$}
\STATE $\overline{\theta}=\vec{\theta}$ \qquad\quad\ \  \COMMENT{this time $(\overline{\theta},\overline{\theta^2},\overline{g},\overline{g^2})$ are vector}
\STATE $\overline{g}=\vec{0}$ \qquad\quad\ \ \COMMENT{initial mean $g$ as zero vector}
\STATE $\forall_i\, \overline{\theta^2}_i=\vec{\theta}_i^2+(\sigma_\theta^0)^2$ \quad\   \COMMENT{some e.g. fixed initial variance}
\STATE $\forall_i\, \overline{g^2}_i=(\sigma_g^0)^2$ \qquad\quad \ \ \  \COMMENT{some e.g. fixed initial variance}
\REPEAT
\STATE $\vec{g}\leftarrow $grad$(\vec{\theta})$ \qquad \COMMENT{find gradient in current position}
\STATE $\overline{\theta}\leftarrow \beta\, \overline{\theta} + (1-\beta) \,\vec{\theta}$ \qquad\COMMENT{as in momentum method}
\STATE $\overline{\theta^2}\leftarrow \beta\, \overline{\theta^2} + (1-\beta)\, \vec{\theta}^{\,2}$ \qquad\ \COMMENT{all coordinate-wise}
\STATE $\overline{g}\leftarrow \beta\, \overline{g} + (1-\beta)\, \vec{g}$
\STATE $\overline{g^2}\leftarrow \beta\, \overline{g^2} + (1-\beta)\, \vec{g}^{\,2}$ \qquad\COMMENT{as in RMSprop~\cite{rmsprop}}
$$\forall_i\ \theta_i\leftarrow \theta_i - \alpha \sqrt{\frac{\overline{\theta^2}_i-\overline{\theta}_i^2}
{\overline{g^2}_i-\overline{g}_i^2}}\,g_i\qquad\qquad\{\textrm{proper step}\}$$
\UNTIL step limit or convergence condition
\end{algorithmic}
}
\end{algorithm}
While we could apply such learning rate to a single adapted direction as previously, thanks to simplicity we can alternatively just model and apply it individually for each parameter $\theta_i$. Such ASGD example is presented in Algorithm \ref{aSGD} as optimization for standard small batch SGD by updating four exponential moving averages $(\overline{\theta}_i,\overline{\theta^2}_i,\overline{g}_i,\overline{g^2}_i)$ separately for each parameter $\theta_i$ - getting a method resembling standard ones like RMSprop~\cite{rmsprop} or ADAM~\cite{adam}, but finally being second order: having potential to minimize parabola in one step, independently for each coordinate.

The details have yet to be optimized based on experiments. Some remarks:

\begin{itemize}
  \item There might be additionally needed some tiny positive $\epsilon$ in denominator of proper step to prevent its zeroing (use $\overline{g^2}_i-\overline{g}_i^2+\epsilon$), but especially for small batches this variance should be relatively large as it also includes variance of gradient estimation (discussed in the next subsection). It might be also worth to consider using negative $\epsilon$ to try to subtract contribution of this noise.
  \item $\sigma_\theta^0$, $\sigma_g^0$ are some initial dispersions (e.g. 1),  $\alpha\,\sigma_\theta^0/ \sigma_g^0$ is initial learning rate - which can be chosen large at the beginning, then it will be adapted from the data in the first few steps.
  \item There might be needed some clipping of learning rate - but rather from below to prevent approaching zero learning rate (e.g. $\max(\sigma_\theta/\sigma_g, \delta)$ for some small $\delta>0$), the next subsection discusses this issue.
  \item The $\sigma_\theta$ and $\sigma_g$ are estimated local dispersions by updating exponential averages of value and squared value like $\overline{\theta}$ and $\overline{\theta^2}$. Instead, alternative approach for the latter might be directly updating variance as central moment, e.g. $\textrm{var}_\theta^{t+1}=\beta' \textrm{var}_\theta^{t} + (1-\beta')\left(\theta^t-\overline{\theta}^t\right)^2$ with not necessarily equal $\beta'$ rate.
\end{itemize}
While $\sqrt{\overline{\theta^2}_i-\overline{\theta}_i^2}$ nominator seems not having been considered in literature before, it is popular to use related $\sqrt{\overline{g^2}}$ denominator started probably by RMSprop~\cite{rmsprop}. Its motivation is rater to strengthen evolution of rare coordinates. For 2nd order modelling it is crucial to also subtract $\overline{g}^2$ to get variance evaluating dispersion, instead of just squared distance from zero.
\subsection{Including noise in learning rate approximations}
In SGD instead of knowing the real gradients (denote them as $\check{g}$), we work on noisy gradients $(g)$ due to finite batch size - we can assume that they are e.g. from Gaussian distribution of standard deviation $\sigma_i$ (which can vary between parameters $i$, generally also time $t$):
\be g_i^t\sim \mathcal{N}(\check{g}_i^t,\sigma_i)\qquad \qquad\sigma_i\propto \frac{1}{\sqrt{\textrm{batch size}}}\ee
This gradient noise is included in the above $\sigma_\theta/\sigma_g$ learning rate estimation, artificially increasing denominator - to remove it we would need to use $\sqrt{(\sigma_g)^2-\sigma^2}$ denominator instead, but it seems difficult to realize, especially when this difference is close to zero.

Such artificially increased denominator can lead to nearly zero learning rate near minimum. It is not necessarily bad, e.g. should give tendency to find wider minima - which are believed to be better from generalization perspective~\cite{flat}. It can be also prevented e.g. by using $\max(\sigma_\theta/\sigma_g, \delta)$ instead for some small $\delta>0$ determining minimal learning rate. However, e.g. when we would like to remain in a varying minimum for some continuously optimized parameters~\cite{adapt}, this error of gradient estimation becomes dominant.

\begin{algorithm}[b!]
\normalsize{
\caption{\textbf{ASGDc}(): using confidence level $\gamma$}
\label{aSGD1}
\begin{algorithmic}
\STATE $\vec{\theta}\in \mathbb{R}^D$ \qquad\quad \COMMENT{initial parameters}
\STATE $\alpha>0$ \qquad\quad\ \ \COMMENT{confidence in parabola model, $\sim 0.5$}
\STATE $\beta\in (0,1)$ \quad\ \ \COMMENT{forgetting rate in averages, $\sim 0.9, 0.99$}
\STATE $\gamma>0$ \qquad\quad \COMMENT{confidence level for $\lambda$ estimation, $\sim 0.1$}
\STATE $\vec{\nu}=\vec{\sigma}^2$\qquad\quad \COMMENT{gradient estimation error, adapted here}
\STATE $\overline{\theta}=\vec{\theta}$ \qquad\quad\ \  \COMMENT{this time $(\overline{\theta},\overline{\theta^2},\overline{g},\overline{g^2})$ are vector}
\STATE $\overline{g}=\vec{0}$ \qquad\quad\ \ \COMMENT{initial mean $g$ as zero vector}
\STATE $\overline{\theta^2}=\vec{\theta}^2+(\sigma_\theta^0)^2$ \qquad\ \COMMENT{some e.g. fixed initial variance}
\STATE $\overline{g\theta}=\vec{0}$ \qquad\quad \ \ \  \COMMENT{for covariance estimation}
\REPEAT
\STATE $\vec{g}\leftarrow $grad$(\vec{\theta})$ \qquad \COMMENT{find gradient in current position}
\STATE $\overline{\theta}\leftarrow \beta\, \overline{\theta} + (1-\beta) \,\vec{\theta}$ \qquad\COMMENT{as in momentum method}
\STATE $\overline{\theta^2}\leftarrow \beta\, \overline{\theta^2} + (1-\beta)\, \vec{\theta}^{\,2}$ \quad\ \ \COMMENT{coordinate-wise square}
\STATE $\overline{g}\leftarrow \beta\, \overline{g} + (1-\beta)\, \vec{g}$
\STATE $\overline{g\theta}\leftarrow \beta\, \overline{g\theta} + (1-\beta)\, \vec{g}\,\vec{\theta}$ \qquad\COMMENT{all coordinate-wise}
\STATE $\vec{\lambda} = \left(\overline{g\theta}-\overline{g}\,\overline{\theta}\right)/\left(\overline{\theta^2}-\overline{\theta}^2 \right)$\qquad \COMMENT{linear coefficients}
\STATE $\vec{\nu}=\beta\,\vec{\nu}+(1-\beta)\left(\vec{g}-\overline{g} - \vec{\lambda}(\vec{\theta}-\overline{\theta})    \right)^2 $\qquad \COMMENT{$\nu=\sigma^2$}
$$\vec{\theta}\leftarrow \vec{\theta}- \alpha\left(\left|\vec{\lambda}\right|+\gamma\sqrt{\frac{\vec{\nu}}
{\overline{\theta^2}-\overline{\theta}^2 }}\, \right)^{-1} \vec{g}\qquad\{\textrm{proper step}\} $$
\UNTIL step limit or convergence condition
\end{algorithmic}
}
\end{algorithm}

Let us discuss a more sophisticated way for estimating learning rate from linear regression of gradients - including this noise from gradient estimation.
Let us derive variance of $\lambda$ assuming variance of $g$ is $\sigma^2$ for weighted case. For simplicity let us shift to the center for a moment: assume $\overline{\theta}=0$, hence $\lambda=\overline{g\theta}/\overline{\theta^2}$:
\be \textrm{var}(\lambda)=\textrm{var}\left(\frac{\sum_i w_i\,g_i\, \theta_i }{\overline{\theta^2}}\right)=\frac{\sigma^2}{\overline{\theta^2}^2} \sum_i w_i^2 \theta_i^2 \label{var1}\ee

\noindent In the used exponential moving average $w_i=(1-\beta)\beta^i$ for $i\geq 0$, leading to $\sum_{i=0}^\infty (w_i)^2=(1-\beta)/(1+\beta)$ - we can approximate sum in (\ref{var1}) with  its product with $\overline{\theta^2}$ (assuming variance of $\theta$ is locally constant, could be also calculated more accurately), leading to formula for general $\overline{\theta}$:
\be \sqrt{\textrm{var}(\lambda)} \approx \frac{\sigma}{\sqrt{\overline{\theta^2}-\overline{\theta}^2}}\, \sqrt{\frac{1-\beta}{1+\beta}} \ee
We can assume that $\lambda$ is approximately from Gaussian distribution of the above standard deviation. To prevent problematic $\lambda\approx 0$, we can shift it from 0 by $\gamma\, \sigma/\sigma_\theta$ for some $\gamma>0$ determining confidence level in units of standard deviations of Gaussian, multiplied by $\sqrt{(1-\beta)/(1+\beta)}$. Additionally, taking absolute value to repel from saddles, and multiplying by $0<\alpha\leq 1$ parameter describing confidence in parabola model (as the real function is not exactly parabola), we get a looking safe practical approximation of the original $\lambda^{-1}$ learning rate:
\be\eta=\alpha \left(\frac{|\overline{g\theta}-\overline{g}\,\overline{\theta}|}{\overline{\theta^2}-\overline{\theta}^2 }+ \frac{\gamma\,\sigma}{\sqrt{\overline{\theta^2}-\overline{\theta}^2 }}\right)^{-1}\quad\textrm{learning rate}\ee
We additionally need to know $\sigma$, which could be estimated e.g. by looking at variance of gradients in a single point. More sophisticated way is adapting it e.g. by approximating the real gradient using the found linear regression:  $\check{g}=f'(\theta)=\lambda(\theta-p)$ where $p=\overline{\theta}-\lambda^{-1}\overline{g}$, getting:
$$g-\check{g}\approx g-\lambda\left(\theta-(\overline{\theta}-\lambda^{-1}\overline{g})\right)=
g-\overline{g}-\lambda(\theta-\overline{\theta})$$

Algorithm \ref{aSGD1} is example of its realization with such adaptation of $\sigma=\sqrt{\nu}$  gradient estimation error, a simpler way is using fixed $\sigma$ e.g. from gradient variance in one point.

\subsection{Estimation and diagonalization of Hessian}
The $i$-th coordinate of $\sigma_\theta/\sigma_g$ in proper step is estimation of $(H_{ii})^{-1}$, hence this way we approximate Hessian $H$ as diagonal matrix.
We can analogously estimate nondiagonal terms by updating also averages of products of $\theta$ and $g$ pairs (but it neglects sign):
\be H_{ij} \approx \sqrt{\frac{\overline{g_i g_j}-\overline{g}_i \overline{g}_j}
{\overline{\theta_i \theta_j}-\overline{\theta}_i \overline{\theta}_j}}\label{hess}\ee

Performance of  2nd order methods can often be improved if diagonalizing Hessian e.g. in some significant subspace - especially that it turns out that gradient is mainly in $d\approx 10$ dimensional top Hessian eigenspace, which is approximately maintained during training~\cite{subs}.

Hence it seems sufficient to model Hessian in $d\approx 10$ dimensional top eigenspace, which should be pointed by sequence of recent gradients as statistically dominating - such model could be maintained using online PCA~\cite{OPCA} of gradient sequence (with exponential moving average), or simpler approach proposed in \cite{me} - just adding new gradient times some tiny constant to all vectors of the considered basis, additionally maintaining their orthonormality.

Having such subspace defined in a current moment by approximately orthornormal basis $\{\vec{v^i}\}_{i=1..d}$, we can project $\vec{\theta}$ and $\vec{g}$ on this basis, use projections to update the four averages in $d$ dimensions to periodically calculate $d\times d$ Hessian using (\ref{hess}), to rotate the basis for maintaining diagonalized Hessian - analogously as discussed in \cite{me}.

\section{Conclusions and further work}
While first order methods do not have a direct way for choosing step size accordingly to local situation, second order parabola model in current direction can provide such optimal step size. While it could be used in a separate line search, here is suggested to be combined e.g. with momentum method. Thanks to linear regression of gradients, 1) get this information online: continuously adapting to local situation, 2) using only gradients already required for momentum method, 3) in practically negligible cost thanks to operating on averages.

Choosing the details like hyperparameters, which generally could evolve in time, is a difficult experimental problem which will require further work. Approximated $\sigma_\theta/\sigma_g$ learning rate seems a simpler practical 2nd order approach.

The general possibility of combining different optimization approaches seems promising and nearly unexplored, starting e.g. with momentum+SGD hybrid: rare large certain steps interleaved with frequent small noisy steps.

There is popular technique of strengthening underrepresented coordinates e.g. in ADAM~\cite{adam}, which might be worth combining with simple second order methods like discussed. They exploit simple exponential moving averages - here we got motivation for exploring further possible averages.

Getting a successful second order method for $d=1$ dimensional subspace, a natural research direction will be increasing this dimension discussed in $\cite{me}$, e.g. to OGR10d. Choosing promising subspace (worth second order modelling) will require going from momentum method to e.g. online-PCA~\cite{OPCA} of recent gradients. Averages $\overline{\theta}$, $\overline{g}$ become $d$ dimensional vectors, $\overline{g\theta}$ and $\overline{\theta\theta}$ become $d\times d$ dimensional matrices, with estimated Hessian: $H=\left(s\overline{g\theta}-\overline{g}\,\overline{\theta}^T\right)
\left(s\overline{\theta\theta}-\overline{\theta}\,\overline{\theta}^T\right)^{-1}$.

\bibliographystyle{IEEEtran}
\bibliography{cites}
\end{document}